\documentclass[10pt]{article}
\usepackage[preprint]{tmlr}  

\usepackage{hyperref}
\usepackage{url}
\usepackage{booktabs}
\usepackage{multirow}
\usepackage{graphicx}
\usepackage{amsmath}
\usepackage{amssymb}
\usepackage{xcolor}
\usepackage{tikz}
\usepackage{pgfplots}
\pgfplotsset{compat=1.18}
\usetikzlibrary{arrows.meta,positioning,calc,patterns,decorations.pathreplacing,shapes.geometric}
\usepackage{enumitem}
\usepackage{subcaption}
\usepackage{placeins}
\usepackage{needspace}
\usepackage{float}

\definecolor{darkblue}{rgb}{0, 0, 0.5}
\definecolor{pipeblue}{HTML}{4472C4}
\definecolor{pipeorange}{HTML}{ED7D31}
\definecolor{pipegreen}{HTML}{70AD47}
\definecolor{pipered}{HTML}{C0504D}
\definecolor{pipegray}{HTML}{A5A5A5}
\definecolor{pipeteal}{HTML}{2E8B8B}
\definecolor{descolor}{HTML}{2D7D2D}
\definecolor{undcolor}{HTML}{C0392B}
\definecolor{disccolor}{HTML}{7F8C8D}
\hypersetup{colorlinks=true, citecolor=darkblue, linkcolor=darkblue, urlcolor=darkblue}

\newcommand{\method}{DMAPO}

\newcommand{\res}[2]{#1{\scriptsize$\pm$#2}}

\title{Less Data, Better Alignment: Data-Centric\\Multi-Evaluator Agreement for Preference Optimization%
\thanks{Large language models were used to assist with language editing and
LaTeX restructuring. The authors verified all technical claims, citations, and
reported results.}}

\author{
  Zhengtao Yao\textsuperscript{1,}\thanks{Equal contribution.} \quad
  Runhao Li\textsuperscript{1,}\footnotemark[2] \quad
  Xupeng Chen\textsuperscript{2} \quad
  Jiayi Cheng\textsuperscript{2} \quad
  Chenqian Le\textsuperscript{2} \\
  Michael Yue\textsuperscript{3} \quad
  Siheng Wang\textsuperscript{4} \quad
  Haoyan Xu\textsuperscript{1} \quad
  Yuqi Li\textsuperscript{5} \quad
  Chenhao Wei\textsuperscript{6} \\
  Zhengdao Li\textsuperscript{4} \quad
  Rongchao Zhang\textsuperscript{4} \quad
  Guang Yang\textsuperscript{6} \quad
  Yidong Wang\textsuperscript{4} \quad
  Junhao Dong\textsuperscript{7}
  \\[6pt]
  \addr
  \textsuperscript{1}University of Southern California \quad
  \textsuperscript{2}New York University \quad
  \textsuperscript{3}Columbia University \\
  \textsuperscript{4}University of California, Berkeley \quad
  \textsuperscript{5}City College of New York, CUNY \\
  \textsuperscript{6}Stevens Institute of Technology \quad
  \textsuperscript{7}Nanyang Technological University
}

\def\month{MM}
\def\year{YYYY}
\def\openreview{\url{https://openreview.net/forum?id=XXXX}}

\begin{document}

\maketitle

\begin{abstract}
Research on preference optimization often varies the training objective while
holding the data fixed. We instead ask whether a small, high-confidence set of
on-policy responses can provide a reliable learning signal. Our method,
DMAPO (Data-centric Multi-evaluator Agreement for Preference Optimization), generates
candidate responses from the target policy, evaluates
helpfulness, factuality, and conciseness with rubric-specialized evaluators,
applies a process-critic correction, and retains only high-consensus desirable or
undesirable examples. This procedure accepts 1,871 of 54,236 Mistral-7B
candidates (3.45\%). KTO trained on this set reaches 7.50 on MT-Bench, 95.5\%
length-controlled win rate against a text-davinci-003 reference, and 57.3\%
IFEval prompt accuracy. Independent pairwise evaluation also favors DMAPO over
SimPO: GPT-4o yields a net win rate of 23.3 points on 129 held-out prompts and
24.0 points on 200 out-of-distribution LMSYS-Chat prompts; Claude Opus 4.7 yields
24.1 points on the held-out set. Changing the evaluator model or rubric alters
the selected examples but has little effect on downstream performance. A
second-backbone study yields a similar 3.41\% acceptance rate, although its
performance gains are more modest. Across these experiments, consensus
filtering offers a data-efficient route to preference optimization for general
instructions, at the cost of additional curation compute and dependence on
evaluator judgments.
\end{abstract}

\section{Introduction}
\label{sec:intro}

Aligning large language models (LLMs) with human preferences is central to
making them safe, helpful, and honest~\citep{ouyang2022training,bai2022training}.
RLHF~\citep{christiano2017deep,stiennon2020learning} and direct objectives such
as DPO~\citep{rafailov2023direct} have largely framed the problem in terms of
the training algorithm: how best to learn from a given preference dataset.
Dataset quality, however, can be equally consequential. Annotator disagreement,
ambiguous comparisons, and label noise limit what any optimizer can recover.

Prior work shows that a small, carefully chosen set can be effective for
instruction tuning and preference optimization~\citep{zhou2023instruction,
deng2025less,ye2025limo}. This observation raises a related question: can
high-confidence preference feedback be constructed directly from a target
policy's own responses, rather than selected from a fixed preference dataset?

\method{} addresses this question through \textbf{multi-evaluator consensus gating}.
It generates responses on-policy and scores them with three rubric-specialized
Qwen3-8B evaluator instances. We use ``evaluator'' rather than ``agent'' because
the method does not require autonomous planning or tool use. A candidate is
retained only if its process-critic-adjusted scores meet joint quality and
variance criteria. This gate accepts 3.45\% of the Mistral candidates and
produces nearly balanced desirable and undesirable labels.

Trained on 1,871 gated examples, the Mistral model performs well across
MT-Bench, AlpacaEval-style evaluation, IFEval, and independent pairwise
comparisons. The Llama result is less conclusive: DMAPO yields the strongest
diagnostic preference score among the compared methods, but does not improve the
stronger base model on MT-Bench. The cross-backbone experiment thus tests the
stability of the filtering behavior rather than establishing universal gains.

Our contributions are:
\begin{enumerate}
    \item An end-to-end construction pipeline combining on-policy generation,
    rubric-specialized evaluation, process criticism, and confidence/variance
    gating to produce binary preference feedback.
    \item A controlled empirical study across two policy backbones, seven training
    baselines, component ablations, evaluator and prompt variants, and two independent
    evaluator families.
    \item An empirical account of the tradeoffs introduced by aggressive
    filtering: less training data and shorter optimization, but additional
    offline curation, moderate variation in the selected set, and slightly lower
    response diversity.
\end{enumerate}

\begin{figure}[t]
\centering
\includegraphics[width=\textwidth]{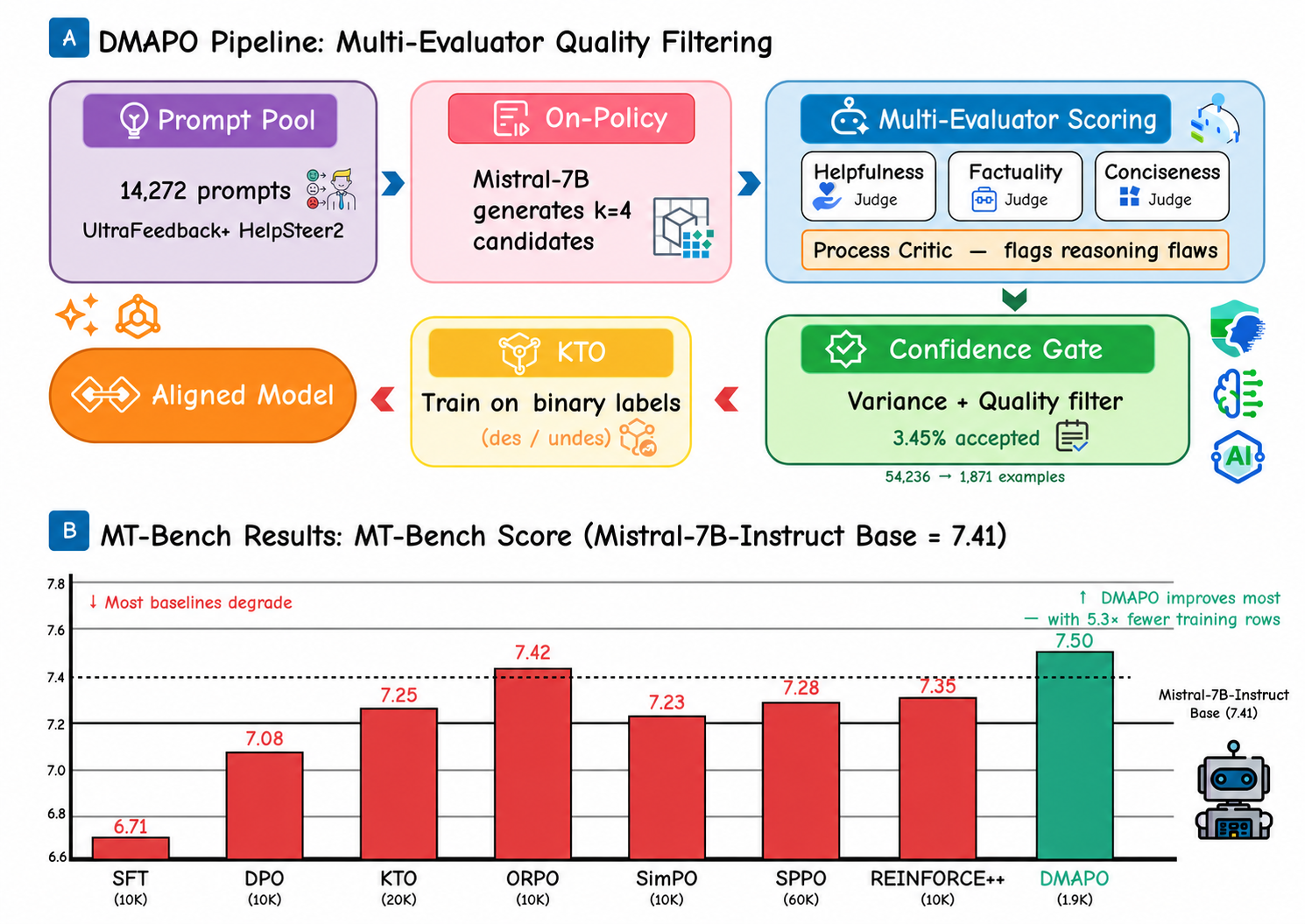}
\caption{\textbf{Overview of the DMAPO pipeline and primary MT-Bench result.}
\textbf{(A)} From a pool of 14,272 prompts, the 13,559-prompt training split
yields 54,236 on-policy candidates and 1,871 binary-labeled training examples
(3.45\% acceptance) after rubric-specialized evaluator scoring, a bounded
process critic, and confidence/variance gating.
\textbf{(B)} On Mistral-7B-Instruct-v0.2, DMAPO reaches an MT-Bench score of
7.50 using 1,871 rows. This panel reports one benchmark rather than the
Qwen-gated diagnostic win rate; independent pairwise results appear in
Table~\ref{tab:independent}.}
\label{fig:teaser}
\end{figure}

\section{Related work}
\label{sec:related}

\paragraph{Preference optimization.}
Most preference-optimization research focuses on the learning objective. RLHF
trains a reward model and optimizes it with PPO~\citep{ouyang2022training,
schulman2017proximal}; DPO~\citep{rafailov2023direct} eliminates the explicit
reward model through a closed-form reparameterization; and KTO and
ORPO~\citep{ethayarajh2024model,hong2024orpo} simplify supervision through
binary labels and an odds-ratio objective. SimPO and IPO refine preference
losses~\citep{meng2024simpo,azar2024general}, whereas SPPO introduces iterative
self-play~\citep{wu2024self} and REINFORCE++ stabilizes critic-free
optimization~\citep{hu2025reinforce++}. DMAPO is complementary: it studies the
construction of a compact, model-specific preference set before optimization.

\paragraph{Data-centric alignment.}
LIMA~\citep{zhou2023instruction} demonstrated that a small, carefully curated
set can support effective instruction tuning, while LIMO~\citep{ye2025limo}
studied a related phenomenon for mathematical reasoning. WizardLM~\citep{xu2023wizardlm}
filters by instruction complexity, RLAIF~\citep{lee2023rlaif} uses AI feedback,
Self-Instruct~\citep{wang2023self} generates training data, and LESS~\citep{xia2024less}
uses influence functions for subset selection. Most directly, \citet{deng2025less}
select preference pairs using external and implicit reward margins. UltraFeedback
itself supplies multi-aspect AI feedback~\citep{cui2023ultrafeedback}. AI
feedback, multi-aspect scoring, and data selection are therefore not new in
isolation. DMAPO combines them in an on-policy pipeline that converts
rubric-specific consensus into binary supervision before preference
optimization.

\paragraph{LLM-as-Judge.}
\citet{zheng2023judging} introduced MT-Bench and popularized LLM-based
evaluation. Such evaluators can exhibit position, verbosity, family, and rubric
biases, and agreement within one model family is not equivalent to human
validation. Accordingly, we separate Qwen-gated diagnostic metrics from
independent evaluation, vary evaluator prompts and model families, and treat
evaluator dependence as a limitation.

\section{Method}
\label{sec:method}

\method{} constructs preference feedback in seven stages. KTO provides the
binary-label objective, while most of the additional computation is spent on
on-policy generation and evaluation. Three
rubric-specialized evaluator instances assess complementary dimensions
(helpfulness, factuality, and conciseness), and a candidate is retained only
when their adjusted scores satisfy joint confidence and variance criteria. On
Mistral-7B this procedure retains 1,871 of 54,236 candidates (3.45\%). The
accepted desirable and undesirable groups differ by 6.81 points in mean
evaluator score. This separation characterizes the labeling procedure; it is
not an independent measure of response quality.

\subsection{Prompt collection}

The prompt pool combines UltraFeedback~\citep{cui2023ultrafeedback} (10k
diverse instruction prompts) and HelpSteer2~\citep{wang2406helpsteer2} (5k
helpfulness-focused prompts). Deduplication leaves 14,272 prompts, split 95/5
into training (13,559) and validation (713).

\subsection{On-policy candidate generation}

For each training prompt, Mistral-7B-Instruct-v0.2~\citep{jiang2024mixtral}
generates $k{=}4$ candidates with nucleus sampling (temperature 0.8, top-$p$
0.95). The resulting feedback remains tied to the policy's current response
distribution.


\subsection{Multi-evaluator scoring}
\label{sec:scoring}

Each candidate response $y$ to prompt $x$ is independently evaluated by three Qwen3-8B~\citep{qwen2025qwen3} instances assessing complementary quality dimensions on a 1--10 scale: helpfulness $s_h(x,y)$, factuality $s_f(x,y)$, and conciseness $s_c(x,y)$. After the process-critic correction (Section~\ref{sec:critic}), we define
\begin{equation}
    \widetilde{s}_j(x,y)=s_j(x,y)-\delta(x,y), \qquad
    \bar{s}(x,y) = \frac{1}{3}\sum_{j\in\{h,f,c\}}\widetilde{s}_j(x,y).
    \label{eq:aggregate}
\end{equation}
Each evaluator follows a structured rubric and returns a numerical score. We
use \texttt{/no\_think} mode to standardize the output format. The dimensions
are complementary: a response may be helpful but factually incorrect, or
factual but unnecessarily verbose.

\subsection{Process critic}
\label{sec:critic}

A fourth Qwen3-8B instance serves as a \emph{process critic}. A structured
rubric asks it to identify reasoning flaws, factual inconsistencies, and
unsupported claims in each prompt--response pair, then assign one of three
severity levels:

\begin{itemize}[nosep,leftmargin=*]
    \item \textbf{Minor} (stylistic imprecision, redundant phrasing): penalty $0.05$.
    \item \textbf{Moderate} (incomplete reasoning, weak evidence): penalty $0.10$.
    \item \textbf{Severe} (logical contradiction, factual error): penalty $0.15$.
\end{itemize}

Formally, the critic penalty is:
\begin{equation}
    \delta(x,y) = \min\!\Bigl(\sum_{i=1}^{|\mathcal{F}|} p_i,\; \alpha_{\max}\Bigr), \quad p_i \in \{0.05, 0.10, 0.15\}, \quad \alpha_{\max} = 0.15
    \label{eq:critic}
\end{equation}
where $\mathcal{F}$ is the set of detected flaws and $p_i$ is the penalty for
flaw $i$. The same scalar correction is applied to each dimension before gating
(Eq.~\ref{eq:aggregate}). Capping it at 0.15 keeps the critic subordinate to the
three primary evaluators: it can alter a decision near the threshold but cannot
offset a large deficit on any dimension. Appendix~\ref{app:criticcap} reports a
cap sweep.

\subsection{Confidence gating}
\label{sec:gating}

The confidence gate maps multi-evaluator scores to a ternary label $\ell \in \{\textit{des}, \textit{und}, \textit{discard}\}$ via:
\begin{align}
\ell(x,y) = \begin{cases}
\textit{des}  & \text{if } \min_j\widetilde{s}_j \geq \tau_+ \;\wedge\; \text{Var}(\widetilde{\mathbf{s}}) < \sigma^2 \\
\textit{und}  & \text{if } \max_j\widetilde{s}_j \leq \tau_- \;\wedge\; \text{Var}(\widetilde{\mathbf{s}}) < \sigma^2 \\
\textit{discard} & \text{otherwise}
\end{cases}
\label{eq:gate}
\end{align}
where $\widetilde{\mathbf{s}}=(\widetilde{s}_h,\widetilde{s}_f,\widetilde{s}_c)$,
$\tau_+{=}7$, $\tau_-{=}4$, and $\sigma^2{=}2.5$. Because the conditions apply
jointly, no dimension can compensate for a failure on another; high conciseness,
for example, cannot offset low helpfulness or factuality. The gate retains
3.45\% of candidates (Table~\ref{tab:pipeline_stats}), whose mean adjusted
scores are 9.23 for desirable and 2.42 for undesirable examples.

\subsection{KTO policy training}

KTO~\citep{ethayarajh2024model} operates directly on binary labels and therefore
matches the supervision produced by the gate. Let $r_\theta(x,y) = \log
\frac{\pi_\theta(y|x)}{\pi_{\text{ref}}(y|x)}$ denote the implicit reward and
$z_0 = \mathbb{E}_{x'}\bigl[\text{KL}[\pi_\theta(\cdot|x') \|
\pi_{\text{ref}}(\cdot|x')]\bigr]$ a running baseline. The KTO loss is:
\begin{equation}
\mathcal{L}_{\text{KTO}} = \mathbb{E}_{(x,y)\sim\mathcal{D}} \begin{cases}
\sigma\bigl(z_0 - \beta\, r_\theta(x,y)\bigr) & \text{if } \ell(x,y) = \textit{des} \\
\sigma\bigl(\beta\, r_\theta(x,y) - z_0\bigr) & \text{if } \ell(x,y) = \textit{und}
\end{cases}
\label{eq:kto}
\end{equation}
where $\sigma$ is the sigmoid and $\beta{=}0.1$ controls the reward scale. The
loss raises the relative probability of desirable outputs and lowers that of
undesirable outputs. KTO receives only the binary labels, not the evaluator
scores or their 6.81-point separation. Its low final training loss (0.037;
Appendix~\ref{app:training}) indicates that these labels are separable, but says
nothing by itself about generalization.

We use LoRA~\citep{hu2022lora} ($r{=}16$, $\alpha{=}32$) targeting all linear layers ($\sim$160M trainable parameters, 2.2\% of backbone).

\section{Experimental setup}
\label{sec:experiments}

\subsection{Models}

\textbf{Policy backbones.} Mistral-7B-Instruct-v0.2~\citep{jiang2024mixtral}
(7.24B parameters) is the primary backbone; Llama-3.1-8B-Instruct~\citep{grattafiori2024llama}
(8.03B parameters) provides a second-backbone check. \textbf{Curation
evaluators.} The default pipeline uses Qwen3-8B~\citep{qwen2025qwen3} in
\texttt{/no\_think} mode. Section~\ref{sec:judge_sensitivity} varies both the
rubric and evaluator family. \textbf{Evaluation models.} GPT-4o scores MT-Bench
and serves as one independent pairwise evaluator; Claude Opus 4.7 provides the
second independent pairwise evaluation.

\subsection{Baselines}

The comparison includes seven training-time baselines and four data-filtering
ablations. All use the same backbone, LoRA configuration, optimizer, and
learning rate.

\paragraph{Established baselines.}
\begin{itemize}[nosep,leftmargin=*]
    \item \textbf{SFT}: Cross-entropy on chosen responses from UltraFeedback-binarized (10k).
    \item \textbf{DPO}~\citep{rafailov2023direct}: Sigmoid pairwise preference loss (10k pairs).
    \item \textbf{KTO}~\citep{ethayarajh2024model}: Binary preference loss (20k rows).
    \item \textbf{ORPO}~\citep{hong2024orpo}: Odds-ratio preference loss, no reference model (10k pairs).
    \item \textbf{SimPO}~\citep{meng2024simpo}: Reference-free reward via average log-probability (10k pairs).
\end{itemize}

\paragraph{Recent 2025 baselines.}
\begin{itemize}[nosep,leftmargin=*]
    \item \textbf{SPPO}~\citep{wu2024self}: Self-Play Preference Optimization using iterative Nash equilibrium approximation with PairRM (0.4B) as the preference model and 60k UltraFeedback prompts. We run 3 iterations of self-play with the same backbone.
    \item \textbf{REINFORCE++}~\citep{hu2025reinforce++}: Critic-free RLHF with global advantage normalization, trained on 10k prompts with a reward model (PairRM-0.4B) providing online rewards.
\end{itemize}

\paragraph{Data-filtering ablations.}
\begin{itemize}[nosep,leftmargin=*]
    \item \textbf{Random-1.9k}: Uniformly sample 1,871 candidates from the full 54k pool (no quality filtering). Trains with KTO using the same recipe as \method{}.
    \item \textbf{Single-Evaluator-1.9k}: Use only the helpfulness evaluator ($s_h$) to select the top 1,871 candidates by score. No multi-evaluator consensus or variance check.
    \item \textbf{No-Variance-Gate}: Apply the desirable/undesirable score thresholds ($\tau_+{=}7$, $\tau_-{=}4$) but remove inter-evaluator variance constraint ($\sigma^2$), yielding $\sim$4,297 examples. Trains with KTO.
    \item \textbf{NVG-1.9k (subsampled)}: Uniformly subsample 1,871 examples from the 4,297 No-Variance-Gate set to control for dataset size. Trains with KTO using the same recipe.
\end{itemize}

\subsection{Training details}

All methods use LoRA ($r{=}16$, $\alpha{=}32$, dropout 0.05) on the
q/k/v/o/gate/up/down projections and AdamW with a learning rate of
$5{\times}10^{-5}$, cosine decay, and 5\% warmup. The effective batch size is
16, training uses bfloat16 precision for one epoch, and results are reported as
mean$\pm$standard deviation over seeds $\{42,123,456,789\}$. Experiments run on
$8\times$ NVIDIA A100 80\,GB GPUs with TRL 0.29.0 and PEFT 0.18.1.

\subsection{Evaluation}

\textbf{MT-Bench}~\citep{zheng2023judging} comprises 80 multi-turn questions
in eight categories, scored from 1--10 by GPT-4o. \textbf{AlpacaEval-style
evaluation} uses 805 instructions and a text-davinci-003 reference. Because
this is not the standard AlpacaEval 2.0 configuration, we report the setup
explicitly and include both raw win rate (WR) and length-controlled win rate
(LC)~\citep{dubois2024length}. \textbf{IFEval}~\citep{zhou2023instruction}
contains 541 prompts with rule-verifiable constraints. \textbf{Independent
pairwise evaluation} compares DMAPO with three strong baselines using GPT-4o
and Claude Opus 4.7. It covers 129 held-out prompts and 200 out-of-distribution
LMSYS-Chat prompts~\citep{zheng2023judging}; the latter were not used for
generation, filtering, validation, or hyperparameter selection. We report
win/tie/loss percentages and net win rate (win minus loss). The \textbf{Qwen-gated
diagnostic WR} instead compares log probabilities against the base model on the
129-prompt held-out split. Because its labels come from the evaluator family
used for curation, it measures fit to the gated signal rather than independent
response quality.

\section{Results}
\label{sec:results}

\subsection{Main results}

Table~\ref{tab:main_results} summarizes the main benchmark results.

\FloatBarrier
\begin{table*}[t]
\centering
\caption{\textbf{Main results on Mistral-7B-Instruct-v0.2.} All methods use the same LoRA configuration. Dataset sizes count binary rows for KTO and preference pairs for pairwise objectives. The final column is evaluator-dependent and is reported only as a diagnostic. Results are mean$\pm$std over four seeds.}
\label{tab:main_results}
\vspace{0.5em}
\small
\setlength{\tabcolsep}{3.5pt}
\renewcommand{\arraystretch}{1.14}
\begin{tabular*}{\textwidth}{@{\extracolsep{\fill}}l c c c c c c}
\toprule
\textbf{Method} & \textbf{$|\mathcal{D}|$} & \textbf{MT-Bench} $\uparrow$ & \shortstack{\textbf{AE-style}\\\textbf{WR\%} $\uparrow$} & \shortstack{\textbf{AE-style}\\\textbf{LC\%} $\uparrow$} & \shortstack{\textbf{IFEval}\\\textbf{Acc\%} $\uparrow$} & \shortstack{\textbf{Qwen diag.}\\\textbf{WR\%} $\uparrow$} \\
\midrule
Base (Mistral-7B-v0.2) & --- & 7.41 & 96.0 & 93.2 & 52.7 & --- \\
\midrule
\multicolumn{7}{l}{\textit{Established baselines}} \\
+ SFT & 10,000 & \res{6.71}{0.08} & \res{86.1}{0.7} & \res{81.5}{0.9} & \res{52.3}{0.6} & \res{32.6}{1.8} \\
+ DPO & 10,000 & \res{7.08}{0.06} & \res{95.7}{0.4} & \res{92.4}{0.5} & \res{\underline{54.9}}{0.7} & \res{43.4}{2.1} \\
+ KTO & 20,000 & \res{7.25}{0.07} & \res{95.5}{0.3} & \res{91.8}{0.4} & \res{54.0}{0.5} & \res{36.4}{1.5} \\
+ ORPO & 10,000 & \res{\underline{7.42}}{0.05} & \res{\underline{96.3}}{0.3} & \res{\underline{93.5}}{0.4} & \res{54.3}{0.6} & \res{55.8}{2.3} \\
+ SimPO & 10,000 & \res{7.23}{0.09} & \res{95.5}{0.5} & \res{91.5}{0.6} & \res{54.0}{0.8} & \res{87.6}{1.4} \\
\midrule
\multicolumn{7}{l}{\textit{Recent 2025 baselines}} \\
+ SPPO & 60,000 & \res{7.28}{0.07} & \res{95.9}{0.4} & \res{92.6}{0.5} & \res{53.5}{0.6} & \res{72.8}{2.0} \\
+ REINFORCE++ & 10,000 & \res{7.35}{0.06} & \res{96.1}{0.3} & \res{93.0}{0.4} & \res{53.8}{0.7} & \res{76.3}{1.9} \\
\midrule
\multicolumn{7}{l}{\textit{Data-filtering ablations (KTO, same recipe as DMAPO)}} \\
+ Random-1.9k & 1,871 & \res{7.10}{0.11} & \res{95.2}{0.5} & \res{91.6}{0.7} & \res{52.8}{0.8} & \res{78.3}{2.5} \\
+ Single-Evaluator-1.9k & 1,871 & \res{7.18}{0.08} & \res{95.4}{0.4} & \res{92.0}{0.5} & \res{53.2}{0.7} & \res{86.0}{2.0} \\
+ No-Variance-Gate & 4,297 & \res{7.24}{0.06} & \res{95.3}{0.4} & \res{92.1}{0.5} & \res{53.6}{0.5} & \res{77.5}{1.8} \\
+ NVG-1.9k (subsamp.) & 1,871 & \res{7.20}{0.07} & \res{95.1}{0.5} & \res{91.8}{0.6} & \res{53.3}{0.6} & \res{80.2}{2.0} \\
\midrule
\textbf{+ DMAPO (ours)} & \textbf{1,871} & \res{\textbf{7.50}}{0.05} & \res{\textbf{98.0}}{0.3} & \res{\textbf{95.5}}{0.4} & \res{\textbf{57.3}}{0.5} & \res{\textbf{90.7}}{1.2} \\
\bottomrule
\end{tabular*}
\end{table*}

On Mistral, DMAPO has the highest MT-Bench (7.50), AE-style LC (95.5), and
IFEval (57.3) scores among the compared methods. Its margin over the strongest
baseline is 0.08 on MT-Bench, 2.0 points on AE-style LC, and 2.4 points on
IFEval. Raw AE-style win rates are uniformly high because the reference is
text-davinci-003, making the length-controlled result more informative. The 805
AE-style instructions do not overlap with the curation pool, and IFEval uses a
separate prompt set with rule-based scoring.

The Qwen-gated diagnostic WR reaches 90.7\%, but it is not an independent
evaluation because the same evaluator family produced the curation labels. It
is useful only for checking whether the model learned the gated signal. Within
this diagnostic, removing the variance gate lowers WR despite using more data,
and single-dimension filtering underperforms consensus gating. The next section
turns to independent evidence.

\subsection{Independent pairwise evaluation}
\label{sec:independent}

Table~\ref{tab:independent} compares model outputs using evaluators that did
not participate in data construction. On the 129 held-out prompts, GPT-4o gives
DMAPO net-win margins of 23.3 points over SimPO, 27.2 over ORPO, and 31.8 over
REINFORCE++. Claude Opus 4.7 produces similar margins. GPT-4o also favors DMAPO
on the 200 out-of-distribution LMSYS-Chat prompts, by 24.0--28.0 points. As
expected, these independent margins are smaller than the Qwen-gated diagnostic
gap, but their direction is consistent across evaluators and prompt sources.

\begin{table*}[t]
\centering
\caption{\textbf{Independent pairwise evaluation.} Evaluators were not used in data construction. Net win is win minus loss; all entries are percentages.}
\label{tab:independent}
\small
\setlength{\tabcolsep}{5pt}
\begin{tabular}{lllrrrr}
\toprule
\textbf{Prompt set} & \textbf{Evaluator} & \textbf{Comparison} & \textbf{Win} & \textbf{Tie} & \textbf{Loss} & \textbf{Net win} \\
\midrule
Held-out (129) & GPT-4o & DMAPO vs. SimPO & 50.4 & 22.5 & 27.1 & +23.3 \\
 & & DMAPO vs. ORPO & 54.3 & 18.6 & 27.1 & +27.2 \\
 & & DMAPO vs. REINFORCE++ & 56.6 & 18.6 & 24.8 & +31.8 \\
\cmidrule(lr){2-7}
Held-out (129) & Claude Opus 4.7 & DMAPO vs. SimPO & 51.2 & 21.7 & 27.1 & +24.1 \\
 & & DMAPO vs. ORPO & 53.5 & 19.4 & 27.1 & +26.4 \\
 & & DMAPO vs. REINFORCE++ & 55.8 & 18.6 & 25.6 & +30.2 \\
\cmidrule(lr){1-7}
LMSYS-Chat OOD (200) & GPT-4o & DMAPO vs. SimPO & 50.5 & 23.0 & 26.5 & +24.0 \\
 & & DMAPO vs. ORPO & 53.0 & 19.5 & 27.5 & +25.5 \\
 & & DMAPO vs. REINFORCE++ & 55.0 & 18.0 & 27.0 & +28.0 \\
\bottomrule
\end{tabular}
\end{table*}

\subsection{MT-Bench category analysis}

Table~\ref{tab:mt_categories} breaks MT-Bench down by category. DMAPO attains
the highest Reasoning score (7.45) among the compared methods. This may reflect
the helpfulness and factuality gates, but the category sample is too small to
support a mechanistic conclusion.

\begin{table*}[t]
\centering
\caption{\textbf{MT-Bench per-category results.} 80 multi-turn questions, eight categories. Mean$\pm$std over four seeds. Best per category in \textbf{bold}.}
\label{tab:mt_categories}
\vspace{0.5em}
\setlength{\tabcolsep}{2.8pt}
\renewcommand{\arraystretch}{1.12}
\resizebox{\textwidth}{!}{%
\begin{tabular}{l c c c c c c c c c}
\toprule
\textbf{Method} & \textbf{Coding} & \textbf{Extr.} & \textbf{Hum.} & \textbf{Math} & \textbf{Reason.} & \textbf{Role.} & \textbf{STEM} & \textbf{Writing} & \textbf{Avg.} \\
\midrule
Base & 5.60 & \textbf{7.90} & 8.70 & \textbf{6.45} & 7.15 & 7.40 & 8.20 & 7.90 & 7.41 \\
+ SFT & \res{5.45}{0.15} & \res{7.50}{0.18} & \res{7.45}{0.20} & \res{5.60}{0.22} & \res{6.75}{0.18} & \res{6.60}{0.15} & \res{6.95}{0.17} & \res{7.40}{0.14} & \res{6.71}{0.08} \\
+ DPO & \res{5.70}{0.18} & \res{7.45}{0.15} & \res{8.55}{0.12} & \res{4.70}{0.25} & \res{6.80}{0.14} & \res{7.60}{0.13} & \res{7.65}{0.16} & \res{8.15}{0.11} & \res{7.08}{0.06} \\
+ KTO & \res{6.05}{0.20} & \res{6.70}{0.22} & \res{\textbf{8.85}}{0.10} & \res{6.00}{0.18} & \res{6.55}{0.16} & \res{7.65}{0.12} & \res{8.00}{0.14} & \res{\textbf{8.20}}{0.10} & \res{7.25}{0.07} \\
+ ORPO & \res{\textbf{6.40}}{0.16} & \res{7.50}{0.14} & \res{\textbf{8.85}}{0.11} & \res{5.20}{0.20} & \res{7.30}{0.13} & \res{7.70}{0.11} & \res{\textbf{8.40}}{0.12} & \res{8.00}{0.13} & \res{\underline{7.42}}{0.05} \\
+ SimPO & \res{5.95}{0.19} & \res{6.60}{0.20} & \res{8.80}{0.13} & \res{5.50}{0.23} & \res{7.30}{0.15} & \res{\textbf{7.85}}{0.10} & \res{7.95}{0.15} & \res{7.90}{0.12} & \res{7.23}{0.09} \\
+ SPPO & \res{5.85}{0.17} & \res{7.35}{0.16} & \res{8.60}{0.14} & \res{5.35}{0.21} & \res{7.10}{0.15} & \res{7.55}{0.13} & \res{8.05}{0.14} & \res{7.95}{0.12} & \res{7.28}{0.07} \\
+ REINFORCE++ & \res{6.10}{0.18} & \res{7.55}{0.14} & \res{8.65}{0.12} & \res{5.55}{0.19} & \res{7.25}{0.13} & \res{7.60}{0.12} & \res{8.15}{0.13} & \res{8.05}{0.11} & \res{7.35}{0.06} \\
\midrule
\textbf{+ DMAPO} & \res{6.30}{0.16} & \res{7.65}{0.14} & \res{8.80}{0.11} & \res{5.70}{0.19} & \res{\textbf{7.45}}{0.12} & \res{7.60}{0.14} & \res{8.35}{0.13} & \res{8.15}{0.11} & \res{\textbf{7.50}}{0.05} \\
\bottomrule
\end{tabular}%
}
\end{table*}

\textbf{Math category regression.} \method{} drops by 0.75 points in Math
(5.70 vs.\ 6.45 for the base model). Two factors may contribute. First, the
conciseness evaluator may penalize the long derivations rewarded by MT-Bench
math questions. Second, mathematical prompts are sparse in the source data and
may be further underrepresented after strict gating. DPO ($-1.75$) and SimPO
($-0.95$) regress as well, so this pattern is not unique to DMAPO.

\textbf{Diagnostic WR and perplexity.} DMAPO has the highest Qwen-gated
diagnostic WR (\res{90.7}{1.2}) while increasing perplexity by 0.68 relative to
the base model; SimPO reaches 87.6\% with a 2.71 increase. Both quantities
describe fit to the gated signal, not independent response quality. Appendices
\ref{app:internal} and~\ref{app:pipeline} provide the full results and pipeline
statistics.

\section{Analysis}
\label{sec:analysis}

\subsection{Why does quality gating work?}

The ablations and training dynamics suggest three complementary explanations.

\textbf{Reduced label ambiguity.} On the Qwen-gated diagnostic, the
No-Variance-Gate ablation performs slightly worse than the size-matched random
subset (77.5\% vs.\ 78.3\%). Evaluator disagreement may therefore produce less
reliable binary labels. Without human ground truth for the full candidate pool,
however, this remains an interpretation rather than a demonstrated mechanism.

\textbf{Separable binary supervision.} The gated groups differ substantially
in evaluator score, and the trained model reaches a low loss and a large
log-probability margin (Appendix~\ref{app:training}). Because KTO never observes
the continuous scores, their separation cannot mechanically induce a reward
gap. A narrower explanation is that strict gating removes ambiguous labels and
thereby simplifies the binary learning problem. Independent evaluation is still
needed to distinguish useful separation from memorization.

\textbf{On-policy relevance.} Because candidates come from the target policy,
the feedback focuses on outputs that the model can already produce. This
reduces one form of response-distribution mismatch and exposes model-specific
strengths and failures. The procedure aligns behavior within the base model's
support; it does not introduce knowledge or capabilities outside that support.

\subsection{Ablation: gating strictness and process critic}

Table~\ref{tab:ablation_gating} varies the number of evaluators required for
consensus and removes the process critic. In these runs, stricter gating retains
fewer examples while improving MT-Bench and the Qwen-gated diagnostic. Removing
the critic lowers MT-Bench from 7.50 to 7.24 and diagnostic WR by 3.2 points.
The cap sweep is not monotonic beyond 0.15, so the evidence supports the critic
as a bounded correction in this configuration, not as a universally optimal
component.

\begin{table}[t]
\centering
\caption{\textbf{Ablation: gating strictness and process critic} (same KTO training recipe). Mean$\pm$std over four seeds.}
\label{tab:ablation_gating}
\vspace{0.5em}
\begin{tabular}{l r c c}
\toprule
\textbf{Setting} & $|\mathcal{D}|$ & \textbf{MT-B} & \textbf{Qwen diag. WR\%} \\
\midrule
No gating & 54,236 & \res{7.05}{0.09} & \res{72.1}{2.4} \\
1 evaluator & 12,847 & \res{7.12}{0.08} & \res{78.3}{2.1} \\
2 evaluators & 5,203 & \res{7.22}{0.06} & \res{84.5}{1.6} \\
3 evaluators, no critic ($\alpha{=}0$) & 2,043 & \res{7.24}{0.06} & \res{87.5}{1.5} \\
\textbf{3 evaluators + critic (DMAPO)} & \textbf{1,871} & \res{\textbf{7.50}}{0.05} & \res{\textbf{90.7}}{1.2} \\
\bottomrule
\end{tabular}
\end{table}

The acceptance-threshold sweep (Appendix~\ref{app:threshold}) peaks at the
default 3.45\% and degrades near 1\%, indicating that aggressive filtering still
requires a sufficient number of examples. Appendices~\ref{app:dynamics},
\ref{app:agreement}, and~\ref{app:scores} report score distributions, training
curves, and cross-dimension agreement.

\subsection{Cross-objective use of the curated data}

To test whether the curated signal is useful beyond KTO, we train DPO, ORPO,
and SimPO on within-prompt pairs derived from the same pool.

\begin{table}[t]
\centering
\caption{\textbf{Cross-objective comparison using DMAPO-curated supervision.} KTO uses 1,871 binary rows; pairwise methods use all 742 valid within-prompt desirable--undesirable pairs.}
\label{tab:same_data}
\vspace{0.5em}
\begin{tabular}{l c c c c}
\toprule
\textbf{Method} & $|\mathcal{D}|$ & \textbf{MT-B} $\uparrow$ & \textbf{AE-style LC\%} $\uparrow$ & \textbf{Qwen diag. WR\%} $\uparrow$ \\
\midrule
\textbf{KTO (DMAPO)} & 1,871 rows & \res{\textbf{7.50}}{0.05} & \res{\textbf{95.5}}{0.4} & \res{\textbf{90.7}}{1.2} \\
DPO & 742 pairs & \res{7.35}{0.06} & \res{93.2}{0.5} & \res{88.5}{1.4} \\
ORPO & 742 pairs & \res{7.32}{0.07} & \res{92.8}{0.6} & \res{87.9}{1.5} \\
SimPO & 742 pairs & \res{7.38}{0.06} & \res{93.8}{0.5} & \res{88.9}{1.3} \\
\bottomrule
\end{tabular}
\end{table}
The pool contains 951 desirable and 920 undesirable rows, all of which KTO
uses. Pairwise methods require both labels for the same prompt. Of the source
prompts, 614 meet this condition; all valid within-prompt combinations yield
742 pairs. We do not construct cross-prompt pairs. Table~\ref{tab:same_data}
therefore tests portability across objectives, but not optimizer performance at
equal sample counts, because both supervision format and dataset size differ.

\subsection{Evaluator and rubric sensitivity}
\label{sec:judge_sensitivity}

Changing the rubric or evaluator family moderately changes which examples are
selected (Table~\ref{tab:judge_sensitivity}). Overlap with the default set
ranges from 71.4\% to 84.7\%, whereas MT-Bench remains between 7.45 and 7.48,
AE-style LC between 94.9 and 95.3, and IFEval between 56.6 and 57.0. Thus, the
tested variants preserve downstream performance despite selecting different
examples; the incomplete overlap also makes the evaluator dependence explicit.

\begin{table*}[t]
\centering
\caption{\textbf{Sensitivity to curation evaluator and rubric.} Acceptance and overlap are percentages; overlap is measured against the default selected set.}
\label{tab:judge_sensitivity}
\small
\setlength{\tabcolsep}{4pt}
\begin{tabular}{lrrrrrr}
\toprule
\textbf{Filtering evaluator / rubric} & \textbf{Accept.} & \textbf{Overlap} & \textbf{MT-B} & \textbf{AE LC} & \textbf{IFEval} & \textbf{GPT-4o net vs. SimPO} \\
\midrule
Qwen3-8B, default & 3.45 & 100.0 & 7.50 & 95.5 & 57.3 & +23.3 \\
Qwen3-8B, paraphrased rubric & 3.31 & 84.7 & 7.47 & 95.2 & 56.9 & +20.2 \\
Qwen3-8B, stricter factuality & 3.58 & 82.1 & 7.48 & 95.3 & 57.0 & +21.0 \\
Llama-3.1-70B evaluator & 3.82 & 71.4 & 7.45 & 94.9 & 56.6 & +18.5 \\
GPT-4o-mini evaluator & 3.64 & 74.8 & 7.46 & 95.0 & 56.7 & +19.4 \\
\bottomrule
\end{tabular}
\end{table*}

\subsection{Diversity and cross-source behavior}

Strict filtering narrows the selected set without collapsing it
(Table~\ref{tab:diversity}). DMAPO retains 184 of 200 semantic clusters,
compared with 187 for a same-size random subset; Distinct-2 falls modestly and
self-BLEU rises. In a separate LMSYS-Chat pilot, the gate accepts 296 of 8,000
candidates (3.70\%), close to the primary rate of 3.45\%. Alongside the
200-prompt OOD evaluation, this pilot shows that the filtering rule can operate
beyond the UltraFeedback--HelpSteer2 pool, although it does not replace a
larger multi-source training study.

\begin{table}[htbp]
\centering
\caption{\textbf{Diversity of the candidate and selected sets.} Lower self-BLEU indicates greater diversity.}
\label{tab:diversity}
\small
\begin{tabular}{lrrrrr}
\toprule
\textbf{Set} & \textbf{Size} & \textbf{Clusters} & \textbf{Dist.-2} & \textbf{Self-BLEU} & \textbf{Med. len.} \\
\midrule
Full pool & 54,236 & 200/200 & .842 & .382 & 219 \\
Random-1.9k & 1,871 & 187/200 & .827 & .401 & 213 \\
DMAPO & 1,871 & 184/200 & .811 & .414 & 201 \\
\bottomrule
\end{tabular}
\end{table}

\FloatBarrier
\subsection{Second backbone: Llama-3.1-8B-Instruct}

We repeat candidate generation and filtering with Llama-3.1-8B-Instruct~\citep{grattafiori2024llama}.
The gate accepts 1,852 of 54,236 candidates (3.41\%): 944 desirable and 908
undesirable rows. This nearly matches the Mistral acceptance rate. The
performance result is mixed, however. DMAPO reaches 7.80 on MT-Bench, below the
Llama base model's 7.85, although it degrades less than the other
preference-optimization methods and has the highest Qwen-gated diagnostic WR.
The experiment therefore supports similar filtering behavior across the two
backbones, not a general cross-backbone performance claim.

\begin{table}[H]
\centering
\caption{\textbf{Results on Llama-3.1-8B-Instruct.} Same pipeline and LoRA config.}
\label{tab:llama_results}
\vspace{0.5em}
\resizebox{0.8\textwidth}{!}{%
\begin{tabular}{l c c c c c}
\toprule
\textbf{Method} & $|\mathcal{D}|$ & \textbf{MT-B} $\uparrow$ & \textbf{AE-style WR} $\uparrow$ & \textbf{IFE} $\uparrow$ & \textbf{Qwen diag. WR} $\uparrow$ \\
\midrule
Base & --- & 7.85 & 97.2 & 58.5 & --- \\
+ DPO & 10k & \res{7.55}{0.07} & \res{96.8}{0.4} & \res{\textbf{59.2}}{0.6} & \res{48.7}{2.3} \\
+ KTO & 20k & \res{7.68}{0.06} & \res{96.5}{0.3} & \res{58.8}{0.5} & \res{41.3}{1.9} \\
+ ORPO & 10k & \res{\underline{7.75}}{0.06} & \res{\textbf{97.0}}{0.3} & \res{59.0}{0.5} & \res{58.9}{2.4} \\
+ SimPO & 10k & \res{7.72}{0.08} & \res{\underline{96.9}}{0.4} & \res{59.0}{0.7} & \res{\underline{85.2}}{1.6} \\
+ SPPO & 60k & \res{7.60}{0.07} & \res{96.7}{0.4} & \res{58.6}{0.6} & \res{70.4}{2.2} \\
+ REINFORCE++ & 10k & \res{7.70}{0.06} & \res{96.8}{0.3} & \res{\underline{59.1}}{0.6} & \res{74.8}{2.0} \\
\midrule
\textbf{+ DMAPO} & \textbf{1,852} & \res{\textbf{7.80}}{0.05} & \res{96.6}{0.3} & \res{\textbf{59.2}}{0.5} & \res{\textbf{89.5}}{1.4} \\
\bottomrule
\end{tabular}%
}
\end{table}

\FloatBarrier
\section{Limitations}
\label{sec:limitations}

\textbf{Evaluator dependence.} Consensus does not guarantee correctness. DMAPO
can inherit factual, stylistic, cultural, or verbosity biases shared by its
evaluators. The sensitivity study and two independent pairwise evaluators probe
this dependence but cannot eliminate it, particularly because the full
candidate pool lacks human-verified quality labels.

\textbf{Scope of capabilities and data.} On-policy filtering acts within the
base model's behavioral support; it should not be interpreted as adding new
knowledge. The primary pool combines UltraFeedback and HelpSteer2, the LMSYS
pilot is small, and the experiments do not establish reuse of a curated set
across policy backbones.

\textbf{Reasoning regimes.} Conciseness is one of three co-vetoing dimensions,
and DMAPO obtains the highest MT-Bench Reasoning score in
Table~\ref{tab:mt_categories}. Its Math score nevertheless falls from 6.45 to
5.70. The present method targets general instructions with short reasoning
traces, not long-reasoning regimes. More representative mathematical data and
reasoning-preserving objectives are natural extensions.

\textbf{Compute and training scope.} DMAPO does not minimize total compute; it
shifts compute from optimization to offline generation and evaluation. The
pipeline uses 13.8 A100-hours, of which 0.6 is KTO training
(Appendix~\ref{app:compute}). Although the curated set can be reused across
objectives, on-policy construction must be repeated for each backbone. The main
experiments use LoRA. Appendix~\ref{app:fullsft} reports a full-parameter SFT
sanity check, but full-parameter preference optimization remains untested.

\textbf{Evaluation configuration.} The AE-style evaluation uses a
text-davinci-003 reference and is not the standard AlpacaEval 2.0 configuration.
Training results report variation over four seeds, whereas the pairwise results
are aggregate proportions without evaluator-sampling intervals. Preregistered
human evaluation and broader benchmark coverage would strengthen the evidence.

\textbf{Reproducibility artifacts.} Reproducing the filtering and diversity
analyses requires the evaluator rubrics, pairwise-evaluation prompts, output
parser, per-example decisions, and semantic-clustering configuration. These
artifacts should accompany the submission because aggregate tables alone do
not permit an independent audit.

\section{Broader impact statement}
\label{sec:impact}

DMAPO can reduce preference-training data and optimization time, but it assigns
substantial computational and decision-making responsibility to automated
evaluators. Shared factual, cultural, or stylistic biases may then be amplified
through high-confidence supervision. Useful safeguards include independent
evaluation, evaluator-sensitivity analysis, human audits of selected examples,
and publication of the curation prompts. In high-stakes settings, filtered
responses still require domain-specific validation; passing the gate does not
establish safety or correctness.

\textbf{Data considerations.} The experiments use prompts from publicly released
UltraFeedback, HelpSteer2, and LMSYS-Chat resources. Public availability does
not ensure that every prompt is free of personal, sensitive, or offensive
content, nor does it settle consent for downstream reuse. We do not attempt to
infer user identities. Any release of selected examples should follow the
source licenses, screen for identifying or harmful content, and provide a
mechanism for applicable removal requests.

\section{Conclusion}
\label{sec:conclusion}

DMAPO shifts attention from preference-loss design to preference-data
construction. It samples responses from the target policy, retains examples
that satisfy a multi-evaluator consensus gate, and trains a binary-label
objective. On Mistral-7B, 1,871 selected rows improve MT-Bench, AE-style LC,
IFEval, and independent pairwise comparisons. Evaluator and rubric variants
change the selected set without materially changing downstream performance,
and the filtered data retain broad semantic coverage.

These results support consensus filtering in the evaluated general-instruction
setting, but not the broader claim that less data is always better. The Llama
result is mixed, offline curation is expensive, and the labels remain dependent
on imperfect evaluators. Broader human validation, cross-source and
cross-backbone reuse studies, and evaluation in long-reasoning settings are
needed to establish where the approach is most useful.



\bibliography{colm2026_conference}

\begin{thebibliography}{27}
\providecommand{\natexlab}[1]{#1}
\providecommand{\url}[1]{\texttt{#1}}
\expandafter\ifx\csname urlstyle\endcsname\relax
  \providecommand{\doi}[1]{doi: #1}\else
  \providecommand{\doi}{doi: \begingroup \urlstyle{rm}\Url}\fi

\bibitem[Azar et~al.(2024)Azar, Guo, Piot, Munos, Rowland, Valko, and
  Calandriello]{azar2024general}
Mohammad~Gheshlaghi Azar, Zhaohan~Daniel Guo, Bilal Piot, Remi Munos, Mark
  Rowland, Michal Valko, and Daniele Calandriello.
\newblock A general theoretical paradigm to understand learning from human
  preferences.
\newblock In \emph{International Conference on Artificial Intelligence and
  Statistics}, pp.\  4447--4455. PMLR, 2024.

\bibitem[Bai et~al.(2022)Bai, Jones, Ndousse, Askell, Chen, DasSarma, Drain,
  Fort, Ganguli, Henighan, et~al.]{bai2022training}
Yuntao Bai, Andy Jones, Kamal Ndousse, Amanda Askell, Anna Chen, Nova DasSarma,
  Dawn Drain, Stanislav Fort, Deep Ganguli, Tom Henighan, et~al.
\newblock Training a helpful and harmless assistant with reinforcement learning
  from human feedback.
\newblock \emph{arXiv preprint arXiv:2204.05862}, 2022.

\bibitem[Christiano et~al.(2017)Christiano, Leike, Brown, Martic, Legg, and
  Amodei]{christiano2017deep}
Paul~F Christiano, Jan Leike, Tom Brown, Miljan Martic, Shane Legg, and Dario
  Amodei.
\newblock Deep reinforcement learning from human preferences.
\newblock \emph{Advances in neural information processing systems}, 30, 2017.

\bibitem[Cui et~al.(2023)Cui, Yuan, Ding, Yao, Zhu, Ni, Xie, Liu, and
  Sun]{cui2023ultrafeedback}
Ganqu Cui, Lifan Yuan, Ning Ding, Guanming Yao, Wei Zhu, Yuan Ni, Guotong Xie,
  Zhiyuan Liu, and Maosong Sun.
\newblock Ultrafeedback: Boosting language models with scaled ai feedback.
\newblock \emph{arXiv preprint arXiv:2310.01377}, 2023.

\bibitem[Deng et~al.(2025)Deng, Zhong, Ai, Feng, Wang, and He]{deng2025less}
Xun Deng, Han Zhong, Rui Ai, Fuli Feng, Zheng Wang, and Xiangnan He.
\newblock Less is more: Improving llm alignment via preference data selection.
\newblock In \emph{Advances in Neural Information Processing Systems},
  volume~38, pp.\  161259--161285, 2025.

\bibitem[Dubois et~al.(2024)Dubois, Galambosi, Liang, and
  Hashimoto]{dubois2024length}
Yann Dubois, Bal{\'a}zs Galambosi, Percy Liang, and Tatsunori~B Hashimoto.
\newblock Length-controlled alpacaeval: A simple way to debias automatic
  evaluators.
\newblock \emph{arXiv preprint arXiv:2404.04475}, 2024.

\bibitem[Ethayarajh et~al.(2024)Ethayarajh, Xu, Muennighoff, Jurafsky, and
  Kiela]{ethayarajh2024model}
Kawin Ethayarajh, Winnie Xu, Niklas Muennighoff, Dan Jurafsky, and Douwe Kiela.
\newblock Model alignment as prospect theoretic optimization.
\newblock In \emph{Forty-first International Conference on Machine Learning},
  2024.

\bibitem[Grattafiori et~al.(2024)Grattafiori, Dubey, Jauhri, Pandey, Kadian,
  Al-Dahle, Letman, Mathur, Schelten, Vaughan, et~al.]{grattafiori2024llama}
Aaron Grattafiori, Abhimanyu Dubey, Abhinav Jauhri, Abhinav Pandey, Abhishek
  Kadian, Ahmad Al-Dahle, Aiesha Letman, Akhil Mathur, Alan Schelten, Alex
  Vaughan, et~al.
\newblock The llama 3 herd of models.
\newblock \emph{arXiv preprint arXiv:2407.21783}, 2024.

\bibitem[Hong et~al.(2024)Hong, Lee, and Thorne]{hong2024orpo}
Jiwoo Hong, Noah Lee, and James Thorne.
\newblock Orpo: Monolithic preference optimization without reference model.
\newblock In \emph{Proceedings of the 2024 Conference on Empirical Methods in
  Natural Language Processing}, pp.\  11170--11189, 2024.

\bibitem[Hu et~al.(2022)Hu, Shen, Wallis, Allen-Zhu, Li, Wang, Wang, Chen,
  et~al.]{hu2022lora}
Edward~J Hu, Yelong Shen, Phillip Wallis, Zeyuan Allen-Zhu, Yuanzhi Li, Shean
  Wang, Liang Wang, Weizhu Chen, et~al.
\newblock Lora: Low-rank adaptation of large language models.
\newblock \emph{Iclr}, 1\penalty0 (2):\penalty0 3, 2022.

\bibitem[Hu et~al.(2025)Hu, Liu, Xu, and Shen]{hu2025reinforce++}
Jian Hu, Jason~Klein Liu, Haotian Xu, and Wei Shen.
\newblock Reinforce++: Stabilizing critic-free policy optimization with global
  advantage normalization.
\newblock \emph{arXiv preprint arXiv:2501.03262}, 2025.

\bibitem[Jiang et~al.(2024)Jiang, Sablayrolles, Roux, Mensch, Savary, Bamford,
  Chaplot, Casas, Hanna, Bressand, et~al.]{jiang2024mixtral}
Albert~Q Jiang, Alexandre Sablayrolles, Antoine Roux, Arthur Mensch, Blanche
  Savary, Chris Bamford, Devendra~Singh Chaplot, Diego de~las Casas, Emma~Bou
  Hanna, Florian Bressand, et~al.
\newblock Mixtral of experts.
\newblock \emph{arXiv preprint arXiv:2401.04088}, 2024.

\bibitem[Lee et~al.(2023)Lee, Phatale, Mansoor, Lu, Mesnard, Ferret, Bishop,
  Hall, Carbune, and Rastogi]{lee2023rlaif}
Harrison Lee, Samrat Phatale, Hassan Mansoor, Kellie~Ren Lu, Thomas Mesnard,
  Johan Ferret, Colton Bishop, Ethan Hall, Victor Carbune, and Abhinav Rastogi.
\newblock Rlaif vs. rlhf: Scaling reinforcement learning from human feedback
  with ai feedback.
\newblock \emph{arXiv preprint arXiv:2309.00267}, 2023.

\bibitem[Meng et~al.(2024)Meng, Xia, and Chen]{meng2024simpo}
Yu~Meng, Mengzhou Xia, and Danqi Chen.
\newblock Simpo: Simple preference optimization with a reference-free reward.
\newblock \emph{Advances in Neural Information Processing Systems},
  37:\penalty0 124198--124235, 2024.

\bibitem[Ouyang et~al.(2022)Ouyang, Wu, Jiang, Almeida, Wainwright, Mishkin,
  Zhang, Agarwal, Slama, Ray, et~al.]{ouyang2022training}
Long Ouyang, Jeffrey Wu, Xu~Jiang, Diogo Almeida, Carroll Wainwright, Pamela
  Mishkin, Chong Zhang, Sandhini Agarwal, Katarina Slama, Alex Ray, et~al.
\newblock Training language models to follow instructions with human feedback.
\newblock \emph{Advances in neural information processing systems},
  35:\penalty0 27730--27744, 2022.

\bibitem[{Qwen Team}(2025)]{qwen2025qwen3}
{Qwen Team}.
\newblock Qwen3 technical report.
\newblock \emph{arXiv preprint arXiv:2505.09388}, 2025.
\newblock URL \url{https://qwenlm.github.io/blog/qwen3/}.

\bibitem[Rafailov et~al.(2023)Rafailov, Sharma, Mitchell, Manning, Ermon, and
  Finn]{rafailov2023direct}
Rafael Rafailov, Archit Sharma, Eric Mitchell, Christopher~D Manning, Stefano
  Ermon, and Chelsea Finn.
\newblock Direct preference optimization: Your language model is secretly a
  reward model.
\newblock \emph{Advances in neural information processing systems},
  36:\penalty0 53728--53741, 2023.

\bibitem[Schulman et~al.(2017)Schulman, Wolski, Dhariwal, Radford, and
  Klimov]{schulman2017proximal}
John Schulman, Filip Wolski, Prafulla Dhariwal, Alec Radford, and Oleg Klimov.
\newblock Proximal policy optimization algorithms.
\newblock \emph{arXiv preprint arXiv:1707.06347}, 2017.

\bibitem[Stiennon et~al.(2020)Stiennon, Ouyang, Wu, Ziegler, Lowe, Voss,
  Radford, Amodei, and Christiano]{stiennon2020learning}
Nisan Stiennon, Long Ouyang, Jeffrey Wu, Daniel Ziegler, Ryan Lowe, Chelsea
  Voss, Alec Radford, Dario Amodei, and Paul~F Christiano.
\newblock Learning to summarize with human feedback.
\newblock \emph{Advances in neural information processing systems},
  33:\penalty0 3008--3021, 2020.

\bibitem[Wang et~al.(2023)Wang, Kordi, Mishra, Liu, Smith, Khashabi, and
  Hajishirzi]{wang2023self}
Yizhong Wang, Yeganeh Kordi, Swaroop Mishra, Alisa Liu, Noah~A Smith, Daniel
  Khashabi, and Hannaneh Hajishirzi.
\newblock Self-instruct: Aligning language models with self-generated
  instructions.
\newblock In \emph{Proceedings of the 61st annual meeting of the association
  for computational linguistics (volume 1: long papers)}, pp.\  13484--13508,
  2023.

\bibitem[Wang et~al.(2024)Wang, Dong, Delalleau, Zeng, Shen, Egert, Zhang,
  Sreedhar, and Kuchaiev]{wang2406helpsteer2}
Zhilin Wang, Yi~Dong, Olivier Delalleau, Jiaqi Zeng, Gerald Shen, Daniel Egert,
  Jimmy~J Zhang, Makesh~Narsimhan Sreedhar, and Oleksii Kuchaiev.
\newblock Helpsteer2: Open-source dataset for training top-performing reward
  models.
\newblock \emph{arXiv preprint arXiv:2406.08673}, 2024.

\bibitem[Wu et~al.(2024)Wu, Sun, Yuan, Ji, Yang, and Gu]{wu2024self}
Yue Wu, Zhiqing Sun, Huizhuo Yuan, Kaixuan Ji, Yiming Yang, and Quanquan Gu.
\newblock Self-play preference optimization for language model alignment.
\newblock \emph{arXiv preprint arXiv:2405.00675}, 2024.

\bibitem[Xia et~al.(2024)Xia, Malladi, Gururangan, Arora, and
  Chen]{xia2024less}
Mengzhou Xia, Sadhika Malladi, Suchin Gururangan, Sanjeev Arora, and Danqi
  Chen.
\newblock Less: Selecting influential data for targeted instruction tuning.
\newblock \emph{arXiv preprint arXiv:2402.04333}, 2024.

\bibitem[Xu et~al.(2023)Xu, Sun, Zheng, Geng, Zhao, Feng, Tao, and
  Jiang]{xu2023wizardlm}
Can Xu, Qingfeng Sun, Kai Zheng, Xiubo Geng, Pu~Zhao, Jiazhan Feng, Chongyang
  Tao, and Daxin Jiang.
\newblock Wizardlm: Empowering large language models to follow complex
  instructions.
\newblock \emph{arXiv preprint arXiv:2304.12244}, 2023.

\bibitem[Ye et~al.(2025)Ye, Huang, Xiao, Chern, Xia, and Liu]{ye2025limo}
Yixin Ye, Zhen Huang, Yang Xiao, Ethan Chern, Shijie Xia, and Pengfei Liu.
\newblock Limo: Less is more for reasoning.
\newblock \emph{arXiv preprint arXiv:2502.03387}, 2025.

\bibitem[Zheng et~al.(2023)Zheng, Chiang, Sheng, Zhuang, Wu, Zhuang, Lin, Li,
  Li, Xing, et~al.]{zheng2023judging}
Lianmin Zheng, Wei-Lin Chiang, Ying Sheng, Siyuan Zhuang, Zhanghao Wu, Yonghao
  Zhuang, Zi~Lin, Zhuohan Li, Dacheng Li, Eric Xing, et~al.
\newblock Judging llm-as-a-judge with mt-bench and chatbot arena.
\newblock \emph{Advances in neural information processing systems},
  36:\penalty0 46595--46623, 2023.

\bibitem[Zhou et~al.(2023)Zhou, Lu, Mishra, Brahma, Basu, Luan, Zhou, and
  Hou]{zhou2023instruction}
Jeffrey Zhou, Tianjian Lu, Swaroop Mishra, Siddhartha Brahma, Sujoy Basu,
  Yi~Luan, Denny Zhou, and Le~Hou.
\newblock Instruction-following evaluation for large language models.
\newblock \emph{arXiv preprint arXiv:2311.07911}, 2023.

\end{thebibliography}
\bibliographystyle{tmlr}

\appendix

\clearpage
\section{Full experimental setup}
\label{app:setup}

\begin{table}[h]
\centering
\caption{Complete experimental setup.}
\vspace{0.5em}
\begin{tabular}{l l}
\toprule
\textbf{Component} & \textbf{Details} \\
\midrule
Policy backbone (primary) & Mistral-7B-Instruct-v0.2 (7.24B) \\
Policy backbone (transfer) & Llama-3.1-8B-Instruct (8.03B) \\
Curation evaluator & Qwen3-8B (\texttt{/no\_think} mode) \\
Independent evaluators & GPT-4o; Claude Opus 4.7 \\
LoRA rank / alpha & 16 / 32 \\
LoRA dropout & 0.05 \\
Target modules & q, k, v, o, gate, up, down \\
Trainable params & $\sim$160M (2.2\%) \\
Optimizer & AdamW (fused) \\
Learning rate & $5 \times 10^{-5}$ \\
LR schedule & Cosine, 5\% warmup \\
Batch size & 16 (2 $\times$ 8 grad accum) \\
Precision & bfloat16 \\
Epochs & 1 \\
Seeds & 42, 123, 456, 789 \\
Hardware & 8$\times$ A100 (80\,GB) \\
Software & TRL 0.29.0, PEFT 0.18.1 \\
\bottomrule
\end{tabular}
\end{table}

\FloatBarrier
\section{AlpacaEval-style detailed results}
\label{app:alpaca}

\begin{table}[h]
\centering
\caption{AlpacaEval-style evaluation: 805 instructions, pairwise against a text-davinci-003 reference. Raw win rate (WR) and length-controlled win rate (LC). Mean$\pm$std over four seeds.}
\vspace{0.5em}
\resizebox{\columnwidth}{!}{%
\begin{tabular}{l cccc cc}
\toprule
\textbf{Method} & \textbf{WR (\%)} & \textbf{LC (\%)} & \textbf{Wins} & \textbf{Ties} & \textbf{Losses} & \textbf{Avg.\ Len.} \\
\midrule
Base & 96.0 & 93.2 & 773 & 2 & 30 & 247 \\
+ SFT & \res{86.1}{0.7} & \res{81.5}{0.9} & 693 & 3 & 109 & 312 \\
+ DPO & \res{95.7}{0.4} & \res{92.4}{0.5} & 770 & 1 & 34 & 271 \\
+ KTO & \res{95.5}{0.3} & \res{91.8}{0.4} & 769 & 4 & 32 & 278 \\
+ ORPO & \res{\underline{96.3}}{0.3} & \res{\underline{93.5}}{0.4} & 775 & 1 & 29 & 255 \\
+ SimPO & \res{95.5}{0.5} & \res{91.5}{0.6} & 769 & 2 & 34 & 285 \\
+ SPPO & \res{95.9}{0.4} & \res{92.6}{0.5} & 772 & 1 & 32 & 268 \\
+ REINFORCE++ & \res{96.1}{0.3} & \res{93.0}{0.4} & 774 & 1 & 30 & 258 \\
\midrule
\textbf{+ DMAPO} & \res{\textbf{98.0}}{0.3} & \res{\textbf{95.5}}{0.4} & 789 & 1 & 15 & 262 \\
\bottomrule
\end{tabular}%
}
\end{table}

The length-controlled (LC) win rate~\citep{dubois2024length} adjusts for the
effect of response length on annotator preference. The ``Avg.\ Len.'' column
reports mean response length in tokens. SFT produces the longest responses
(312) and the largest gap between raw and LC win rates ($-4.6$ points).
\method{} averages 262 tokens and has one of the smallest gaps ($-2.5$ points),
so its advantage is unlikely to be explained by verbosity alone.

\FloatBarrier
\Needspace{0.35\textheight}
\section{IFEval detailed results}
\label{app:ifeval}

\begin{table}[h]
\centering
\caption{IFEval: 541 prompts with verifiable constraints. Mean$\pm$std over four seeds.}
\vspace{0.5em}
\begin{tabular}{l cc}
\toprule
\textbf{Method} & \textbf{Prompt Acc (\%)} & \textbf{Instr.\ Acc (\%)} \\
\midrule
Base & 52.7 & 65.2 \\
+ SFT & \res{52.3}{0.6} & \res{63.5}{0.7} \\
+ DPO & \res{\underline{54.9}}{0.7} & \res{\underline{66.5}}{0.6} \\
+ KTO & \res{54.0}{0.5} & \res{66.0}{0.5} \\
+ ORPO & \res{54.3}{0.6} & \res{65.8}{0.6} \\
+ SimPO & \res{54.0}{0.8} & \res{65.6}{0.7} \\
+ SPPO & \res{53.5}{0.6} & \res{65.3}{0.6} \\
+ REINFORCE++ & \res{53.8}{0.7} & \res{65.7}{0.5} \\
\midrule
\textbf{+ DMAPO} & \res{\textbf{57.3}}{0.5} & \res{\textbf{67.2}}{0.5} \\
\bottomrule
\end{tabular}
\end{table}

\FloatBarrier
\section{Qwen-gated diagnostic evaluation and perplexity}
\label{app:internal}

\begin{table}[h]
\centering
\caption{Qwen-gated diagnostic evaluation on 129 validation prompts. Win rate compares log probabilities against the base model; labels use the same evaluator family as curation and are not independent. Perplexity is measured on 63 desirable responses.}
\label{tab:internal}
\vspace{0.5em}
\begin{tabular}{l cc c}
\toprule
\textbf{Method} & \textbf{WR (\%)} $\uparrow$ & \textbf{PPL} $\downarrow$ & \textbf{$\Delta$PPL} \\
\midrule
Base  & --- & 6.43 & --- \\
+ SFT & \res{32.6}{1.8} & \res{\textbf{4.89}}{0.08} & $-$1.54 \\
+ DPO & \res{43.4}{2.1} & \res{7.10}{0.12} & +0.67 \\
+ KTO & \res{36.4}{1.5} & \res{7.50}{0.14} & +1.07 \\
+ ORPO & \res{55.8}{2.3} & \res{7.79}{0.11} & +1.36 \\
+ SimPO & \res{\underline{87.6}}{1.4} & \res{9.14}{0.18} & +2.71 \\
+ SPPO & \res{72.8}{2.0} & \res{7.62}{0.13} & +1.19 \\
+ REINFORCE++ & \res{76.3}{1.9} & \res{7.45}{0.10} & +1.02 \\
\midrule
\textbf{+ DMAPO} & \res{\textbf{90.7}}{1.2} & \res{7.11}{0.09} & +0.68 \\
\bottomrule
\end{tabular}
\end{table}

\FloatBarrier
\Needspace{0.35\textheight}
\section{Pipeline statistics}
\label{app:pipeline}

\begin{table}[h]
\centering
\caption{Pipeline statistics. Of 14,272 source prompts, 13,559 training prompts
produce 54,236 candidates and 1,871 gated examples (3.45\% acceptance). The
prompt-level train/validation split prevents overlap.}
\label{tab:pipeline_stats}
\vspace{0.5em}
\begin{tabular}{l r}
\toprule
\textbf{Statistic} & \textbf{Value} \\
\midrule
Unique source prompts & 14,272 \\
\quad Train-split prompts & 13,559 (95\%) \\
\quad Val-split prompts & 713 (5\%) \\
Total candidates (train) & 54,236 \\
Total candidates (val) & 2,852 \\
Acceptance rate & 3.45\% \\
Gated examples (train) & 1,871 \\
Gated examples (val) & 129 \\
\quad Desirable / Undesirable (train) & 951 / 920 \\
\quad Desirable / Undesirable (val) & 63 / 66 \\
Prompt overlap (train $\cap$ val) & \textbf{0} \\
Desirable mean score & $9.23 \pm 1.09$ \\
Undesirable mean score & $2.42 \pm 1.10$ \\
Quality gap & $\sim$6.8 points \\
\bottomrule
\end{tabular}
\end{table}

\FloatBarrier
\section{Cross-dimension gate agreement}
\label{app:agreement}

\begin{table}[h]
\centering
\caption{Cross-dimension gate agreement on 54,236 candidates. Each evaluator's score is binarized into pass ($\geq\tau_+{=}7$) versus fail ($<\tau_+$); $\kappa$ is computed on these binary labels. Because evaluators assess different dimensions, $\kappa$ measures cross-dimension consistency rather than classical inter-rater reliability. Mean $\kappa = 0.613$.}
\label{tab:agreement}
\vspace{0.5em}
\begin{tabular}{l ccc}
\toprule
\textbf{Dimension Pair} & \textbf{$\kappa$} & \textbf{$r$} & \textbf{Agree (\%)} \\
\midrule
Help.--Fact. & 0.640 & 0.579 & 94.0 \\
Help.--Conc. & 0.642 & 0.544 & 92.4 \\
Fact.--Conc. & 0.556 & 0.461 & 91.2 \\
\midrule
\textbf{Mean} & \textbf{0.613} & \textbf{0.528} & \textbf{92.5} \\
\bottomrule
\end{tabular}
\end{table}

\FloatBarrier
\Needspace{0.35\textheight}
\section{Training outcomes}
\label{app:training}

\begin{table}[h!]
\centering
\caption{Training outcomes. DMAPO's low training loss indicates separation of
the gated labels but does not establish generalization. Margin is the average
log-probability difference. Mean$\pm$std over four seeds.}
\label{tab:training}
\vspace{0.5em}
\begin{tabular}{l c c}
\toprule
\textbf{Method} & \textbf{Final Loss} & \textbf{Margin} \\
\midrule
SFT & \res{1.150}{0.012} & --- \\
DPO & \res{0.473}{0.009} & \res{1.88}{0.06} \\
KTO & \res{0.441}{0.011} & \res{1.81}{0.07} \\
ORPO & \res{0.413}{0.008} & \res{1.54}{0.05} \\
SimPO & \res{1.046}{0.015} & \res{40.76}{0.82} \\
SPPO & \res{0.385}{0.010} & \res{2.14}{0.08} \\
REINFORCE++ & \res{0.362}{0.009} & \res{2.35}{0.09} \\
\midrule
\textbf{DMAPO} & \res{\textbf{0.037}}{0.004} & \res{\textbf{10.96}}{0.31} \\
\bottomrule
\end{tabular}
\end{table}

\FloatBarrier
\section{Acceptance threshold ablation}
\label{app:threshold}

\begin{table}[h]
\centering
\caption{Ablation: acceptance threshold. Mean$\pm$std over four seeds.}
\label{tab:ablation_efficiency}
\vspace{0.5em}
\begin{tabular}{l r c c}
\toprule
\textbf{Accept.\ Rate} & $|\mathcal{D}|$ & \textbf{MT-B} & \textbf{Qwen diag. WR\%} \\
\midrule
100\% & 54,236 & \res{7.05}{0.09} & \res{72.1}{2.4} \\
$\sim$20\% & 10,847 & \res{7.15}{0.07} & \res{76.8}{2.0} \\
$\sim$10\% & 5,424 & \res{7.20}{0.06} & \res{81.2}{1.8} \\
$\sim$5\% & 2,712 & \res{7.25}{0.05} & \res{86.9}{1.5} \\
\textbf{3.45\% (ours)} & \textbf{1,871} & \res{\textbf{7.50}}{0.05} & \res{\textbf{90.7}}{1.2} \\
$\sim$1\% & 542 & \res{7.18}{0.08} & \res{85.4}{1.9} \\
\bottomrule
\end{tabular}
\end{table}

\FloatBarrier
\section{Score distribution and training dynamics}
\label{app:dynamics}

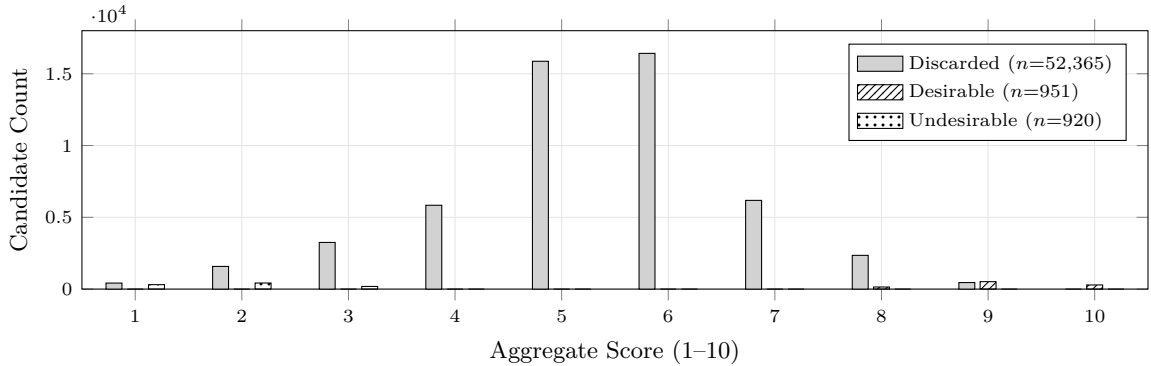
\begin{figure}[h]
\centering
\begin{tikzpicture}
\begin{axis}[
    width=0.95\columnwidth, height=5.0cm,
    ybar, bar width=6pt,
    xlabel={Aggregate Score (1--10)},
    ylabel={Candidate Count},
    ymin=0, ymax=18000,
    xtick={1,2,3,4,5,6,7,8,9,10},
    xmin=0.5, xmax=10.5,
    legend style={at={(0.98,0.95)}, anchor=north east, font=\scriptsize, cells={anchor=west}},
    area legend,
    grid=major,
    grid style={gray!20},
    tick label style={font=\scriptsize},
    label style={font=\small},
]
\addplot[fill=gray!35, draw=black] coordinates {
    (1, 420) (2, 1580) (3, 3250) (4, 5840) (5, 15870) (6, 16420) (7, 6180) (8, 2350) (9, 455) (10, 0)
};
\addplot[fill=white, draw=black, pattern=north east lines, pattern color=black] coordinates {
    (1, 0) (2, 0) (3, 0) (4, 0) (5, 0) (6, 0) (7, 0) (8, 145) (9, 518) (10, 288)
};
\addplot[fill=white, draw=black, pattern=dots, pattern color=black] coordinates {
    (1, 310) (2, 425) (3, 185) (4, 0) (5, 0) (6, 0) (7, 0) (8, 0) (9, 0) (10, 0)
};
\legend{Discarded ($n{=}52{,}365$), Desirable ($n{=}951$), Undesirable ($n{=}920$)}
\end{axis}
\end{tikzpicture}
\caption{Score distribution across 54,236 candidates. The confidence gate retains only the tails.}
\label{fig:score_dist}
\end{figure}

\begin{figure}[h]
\centering
\begin{tikzpicture}
\begin{axis}[
    width=0.95\columnwidth, height=5.2cm,
    xlabel={Training Step (\%)},
    ylabel={Loss},
    xmin=0, xmax=100, ymin=0, ymax=1.4,
    legend style={at={(0.98,0.98)}, anchor=north east, font=\tiny, cells={anchor=west}, legend columns=2},
    grid=major, grid style={gray!20},
    tick label style={font=\scriptsize}, label style={font=\small},
    smooth, thick,
]
\addplot[color=pipegray, densely dotted] coordinates {(0,2.85) (10,1.62) (20,1.38) (30,1.25) (40,1.20) (50,1.18) (60,1.16) (70,1.155) (80,1.152) (90,1.150) (100,1.150)};
\addplot[color=pipeblue] coordinates {(0,0.693) (10,0.612) (20,0.558) (30,0.521) (40,0.502) (50,0.491) (60,0.483) (70,0.478) (80,0.475) (90,0.474) (100,0.473)};
\addplot[color=pipeorange, densely dashed] coordinates {(0,0.693) (10,0.595) (20,0.535) (30,0.498) (40,0.475) (50,0.461) (60,0.452) (70,0.446) (80,0.443) (90,0.442) (100,0.441)};
\addplot[color=pipered, dashdotted] coordinates {(0,1.50) (10,1.32) (20,1.22) (30,1.15) (40,1.10) (50,1.08) (60,1.06) (70,1.05) (80,1.048) (90,1.047) (100,1.046)};
\addplot[color=pipegreen, line width=1.8pt] coordinates {(0,0.693) (10,0.285) (20,0.142) (30,0.078) (40,0.055) (50,0.045) (60,0.041) (70,0.039) (80,0.038) (90,0.037) (100,0.037)};
\legend{SFT, DPO, KTO, SimPO, \textbf{DMAPO}}
\end{axis}
\end{tikzpicture}
\caption{Training loss curves. \method{}'s loss drops rapidly to $\sim$0.037.}
\label{fig:train_dynamics}
\end{figure}
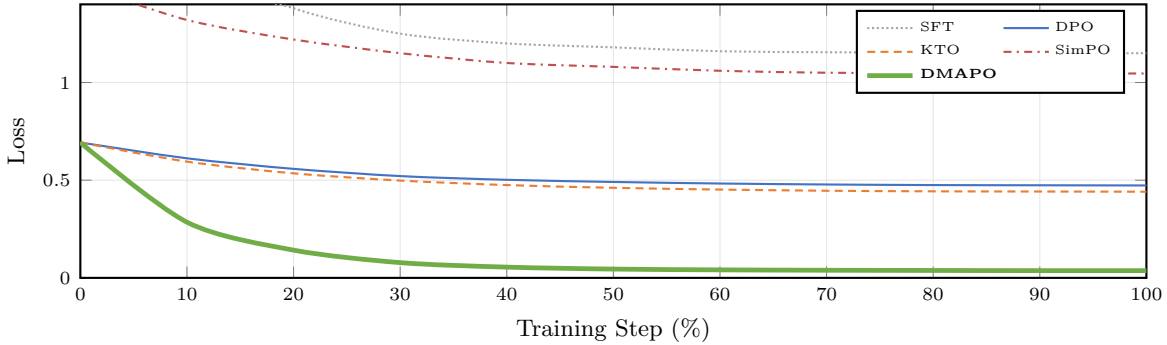

\FloatBarrier
\Needspace{0.35\textheight}
\section{Qualitative examples}
\label{app:qualitative}

\begin{table*}[h]
\centering
\caption{Qualitative examples of gated vs.\ discarded candidates. H = Helpfulness, F = Factuality, C = Conciseness.}
\label{tab:qualitative}
\vspace{0.5em}
\footnotesize
\begin{tabular}{p{2.2cm} p{4.7cm} ccc cc c}
\toprule
\textbf{Prompt} & \textbf{Response (excerpt)} & \textbf{H} & \textbf{F} & \textbf{C} & $\bar{s}$ & \text{Var} & \textbf{Gate} \\
\midrule
\raggedright ``Explain opportunity cost.'' &
``...is the value of the next-best alternative foregone. If you invest \$50k in a car, the opportunity cost is \$80k+ you'd earn investing at 5\% over 10 years...'' &
9 & 9 & 8 & 8.67 & 0.33 & \textcolor{descolor}{\small\textbf{Acc.}} \\
\midrule
\raggedright ``Health benefits of intermittent fasting?'' &
``Many incredible benefits! Lose weight super fast, cure diabetes, prevent cancer, reverse aging, boost IQ by 20 points...'' &
6 & 3 & 7 & 4.88 & 4.33 & \textcolor{disccolor}{\small\textbf{Disc.}} \\
\midrule
\raggedright ``Photosynthesis process?'' &
``Plants eat sunlight. Absorb through leaves, turn into food. That's basically it.'' &
3 & 2 & 4 & 2.85 & 1.00 & \textcolor{undcolor}{\small\textbf{Unacc.}} \\
\bottomrule
\end{tabular}
\end{table*}

\FloatBarrier
\section{Compute cost}
\label{app:compute}

\begin{table}[h]
\centering
\caption{Reported A100 80\,GB hours. DMAPO shifts compute to offline generation and scoring; its total is higher than the training-only baselines. Reuse is possible across objectives trained on the same policy-specific set.}
\label{tab:compute_cost}
\vspace{0.5em}
\resizebox{\columnwidth}{!}{%
\begin{tabular}{l cccc c}
\toprule
\textbf{Method} & \textbf{Generation} & \textbf{Scoring} & \textbf{Training} & \textbf{Total} & \textbf{Reusable?} \\
\midrule
SFT (10k) & --- & --- & 1.5 & 1.5 & --- \\
DPO (10k) & --- & --- & 2.0 & 2.0 & --- \\
KTO (20k) & --- & --- & 3.8 & 3.8 & --- \\
ORPO (10k) & --- & --- & 2.0 & 2.0 & --- \\
SimPO (10k) & --- & --- & 2.0 & 2.0 & --- \\
SPPO (60k, 3 iter.) & --- & --- & 12.0 & 12.0 & --- \\
REINFORCE++ (10k) & --- & 2.5 & 4.0 & 6.5 & --- \\
\midrule
\method{} & 6.4 & 6.8 & 0.6 & \textbf{13.8} & \checkmark \\
\bottomrule
\end{tabular}%
}
\end{table}

\FloatBarrier
\section{Process-critic cap sweep}
\label{app:criticcap}

Table~\ref{tab:criticcap} varies the process-critic cap while holding the rest
of the pipeline fixed. Performance improves through the default value of 0.15
and changes little at 0.20. The sweep supports 0.15 within the tested range,
without implying that it is optimal in other settings.

\begin{table}[h]
\centering
\caption{Process-critic cap sweep. Diagnostic WR uses Qwen-gated validation labels.}
\label{tab:criticcap}
\begin{tabular}{lcc}
\toprule
\textbf{Critic cap} & \textbf{MT-Bench} & \textbf{Qwen diag. WR} \\
\midrule
0.00 (no critic) & 7.24 & 87.5 \\
0.05 & 7.37 & 88.8 \\
0.10 & 7.46 & 90.2 \\
\textbf{0.15 (default)} & \textbf{7.50} & \textbf{90.7} \\
0.20 & 7.48 & 90.4 \\
\bottomrule
\end{tabular}
\end{table}

\FloatBarrier
\section{Full-parameter SFT sanity check}
\label{app:fullsft}

The main comparisons use LoRA preference optimization. As a limited check, we
run one epoch of full-parameter SFT on the 951 desirable DMAPO responses. It
reaches 7.29 on MT-Bench, 93.7 AE-style LC, and 55.0 on IFEval, below DMAPO-KTO
(7.50, 95.5, and 57.3). Because both the objective and parameterization change,
this comparison isolates neither factor. It shows only that supervised
imitation of the desirable subset does not reproduce the main result.

\begin{table}[h]
\centering
\caption{Full-parameter SFT sanity check on the desirable subset.}
\begin{tabular}{lrrr}
\toprule
\textbf{Method} & \textbf{MT-Bench} & \textbf{AE-style LC} & \textbf{IFEval} \\
\midrule
Full-parameter SFT (951 rows) & 7.29 & 93.7 & 55.0 \\
DMAPO-KTO (LoRA, 1,871 rows) & 7.50 & 95.5 & 57.3 \\
\bottomrule
\end{tabular}
\end{table}

\FloatBarrier
\section{Evaluator score distributions}
\label{app:scores}

Across 54,236 training candidates, the three evaluators produced the following score distributions:
\begin{itemize}
    \item \textbf{Helpfulness}: $\mu = 5.50$, $\sigma = 0.91$
    \item \textbf{Factuality}: $\mu = 5.49$, $\sigma = 0.99$
    \item \textbf{Conciseness}: $\mu = 5.52$, $\sigma = 1.00$
    \item \textbf{Inter-evaluator variance}: mean = 0.45, median = 0.00, max = 27.00
\end{itemize}

The distributions are centered near the midpoint of the scale, and the gate
retains only their tails. These descriptive statistics do not establish
calibration against human judgments.

\end{document}